\documentclass[sigconf,nonacm]{acmart}

\setcopyright{none}
\renewcommand\footnotetextcopyrightpermission[1]{}
\usepackage{url}

\usepackage{algorithm}
\usepackage{algpseudocode}
\usepackage{subcaption}
\usepackage{graphicx}
\usepackage{textcomp}
\usepackage{xcolor}
\usepackage{wrapfig}
\usepackage{caption}
\usepackage{multirow}
\usepackage{array}
\usepackage{makecell}
\usepackage[normalem]{ulem}
\usepackage{soul}

\AtBeginDocument{%
  }

\author{Zehao Fan}
\authornote{Both authors contributed equally to this work.}
\affiliation{%
\institution{Rensselaer Polytechnic Institute}
\city{Troy}
\state{NY}
\country{USA}
}

\author{Zhenyu Liu}
\authornotemark[1]  % 复用上面同一个脚注标记（*）
\affiliation{%
\institution{Rensselaer Polytechnic Institute}
\city{Troy}
\state{NY}
\country{USA}
}

\author{Yunzhen Liu}
\affiliation{%
\institution{University of Massachusetts Amherst}
\city{Amherst}
\state{MA}
\country{USA}
}

\author{Yayue Hou}
\affiliation{%
\institution{Rensselaer Polytechnic Institute}
\city{Troy}
\state{NY}
\country{USA}
}

\author{Hadjer Benmeziane}
\affiliation{%
\institution{IBM Research Europe}
\country{Switzerland}
}

\author{Kaoutar El~Maghraoui}
\affiliation{%
\institution{IBM T.~J.~Watson Research Center}
\city{Yorktown Heights}
\state{NY}
\country{USA}
}

\author{Liu Liu}
\affiliation{%
\institution{Rensselaer Polytechnic Institute}
\city{Troy}
\state{NY}
\country{USA}
}

\begin{document}

%%
%% The "title" command has an optional parameter,
%% allowing the author to define a "short title" to be used in page headers.
\title{Context-Aware Mixture-of-Experts Inference on CXL-Enabled GPU–NDP Systems}

%%
%% The "author" command and its associated commands are used to define
%% the authors and their affiliations.
%% Of note is the shared affiliation of the first two authors, and the
%% "authornote" and "authornotemark" commands
%% used to denote shared contribution to the research.

%%
%% The abstract is a short summary of the work to be presented in the
%% article.
\begin{abstract}

 Mixture-of-Experts (MoE) models scale large language models through conditional computation, but inference becomes \emph{memory-bound} once expert weights exceed the capacity of GPU memory. In this case, weights must be offloaded to external memory, and fetching them incurs costly and repeated transfers. We address this by adopting \emph{CXL-attached near-data processing (CXL-NDP)} as the offloading tier to execute cold experts in place, converting expensive parameter movement into cheaper activation movement. Unlike prior GPU–NDP systems that are largely context-agnostic and reactive, we develop a \emph{context-aware} MoE system that uses prefill-stage activation statistics to guide decoding-stage expert placement, dynamically pins hot experts in GPU-side HBM, and maps the remainder to CXL-NDP. To meet NDP's limited compute throughput, we introduce context-aware mixed-precision quantization that allocates per-expert bitwidths (1--4\,bit) based on prefill stage. The resulting MoE inference system overlaps GPU and NDP execution while minimizing cross-device movement. The evaluation on the GPU–NDP system shows that our approach achieves up to 8.7× decoding throughput improvement over state-of-the-art method, while incurring only a 0.13\% average accuracy drop.

\end{abstract}

%%
%% The code below is generated by the tool at http://dl.acm.org/ccs.cfm.
%% Please copy and paste the code instead of the example below.
%%

%%
%% Keywords. The author(s) should pick words that accurately describe
%% the work being presented. Separate the keywords with commas.
%% A "teaser" image appears between the author and affiliation
%% information and the body of the document, and typically spans the
%% page.

%%
%% This command processes the author and affiliation and title
%% information and builds the first part of the formatted document.
\maketitle

\section{Introduction}\label{sec:intro}

Mixture-of-Experts (MoE) models\cite{jiang2024mixtral, liu2024deepseek, yang2025qwen3, shazeer2017outrageously, muennighoff2024olmoe} enable scaling large language models (LLMs) via conditional computation: Each Transformer layer replaces its FFN with a pool of experts and activates only a small subset per token. This sparsity preserves per-token FLOPs while growing parameters, but it typically causes the full model to exceed GPU memory capacity. For example, inference with Mixtral 8$\times$22B~\cite{mistral2024mixtral} in FP16 precision requires approximately 280~GB of memory, far exceeding the memory capacity of a single GPU, and therefore makes inference \emph{memory-bound}: since all experts must remain accessible, naively offloading weights to external memory (e.g., CXL memory) forces frequent parameter transfers over PCIe that dominate latency and reduce GPU utilization. As reported in~\cite{zhang2025daop}, the latency of migrating an expert from the CPU to the GPU can exceed 90\% of the total execution time of a Transformer block, greatly surpassing both expert and non-expert computation.

% \begin{wrapfigure}{r}{0.27\textwidth} % r=右侧，l=左侧，宽度自调
%     \vspace{-10pt}
%     \centering
%     \includegraphics[width=0.27\textwidth]{Figure/pan-Zehao.pdf}
%     \caption{MoE inference time breakdown.}
%     \label{fig:mixtral_offload_profile}
%     \vspace{-5pt}

% \end{wrapfigure}

To overcome this bottleneck, recent work has explored heterogeneous systems that couple GPUs with near-data processing (NDP) devices~\cite{kim2024monde, wu2025pimoe, huang2025hd, yun2024duplex}. Among these, CXL-attached memory with near-data processing (CXL-NDP) provides large-capacity DDR-class memory and high internal bandwidth. These devices can execute expert computation near memory, converting large parameter movement into small activation movement, and they support much larger MoE models at lower cost, making them a practical and promising solution.

\begin{figure*}[t]
  \centering    
  \includegraphics[width=0.99\textwidth]{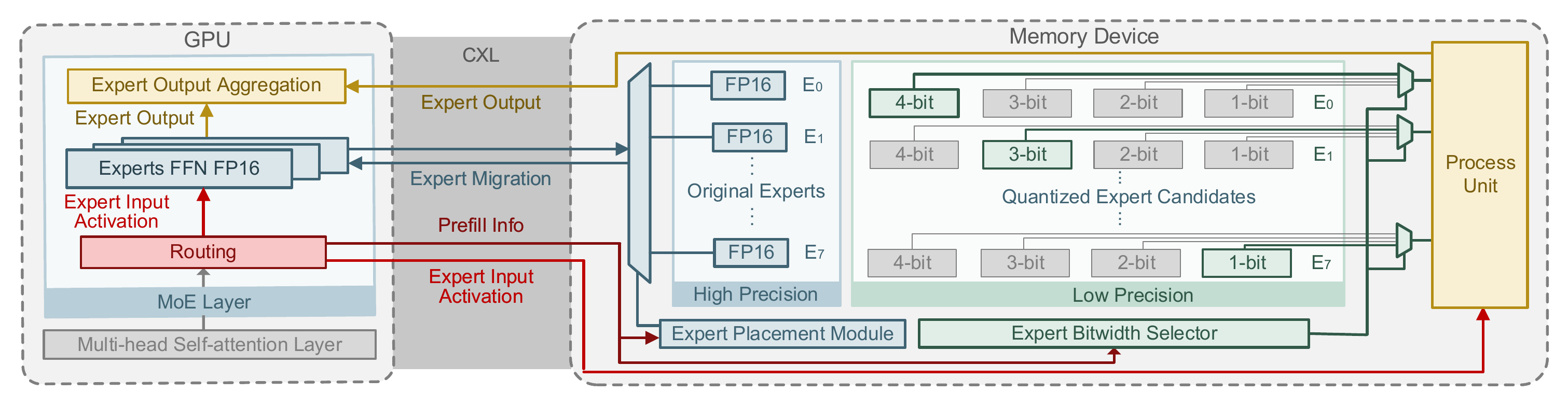}
  \caption{System overview. During MoE inference, prefill-stage expert activation statistics are collected and fed to two modules: the Expert Placement Module, which runs once per sequence to determine an efficient GPU/NDP expert mapping; the Expert Bitwidth Selector, which uses the same statistics to assign per-expert quantization bitwidths on the NDP device, improving system performance while reducing accuracy loss.}
  % 1) the GPU handles the prefill stage and hot experts; 2) Prefill-guided context-aware expert placement, i.e., migrate hot experts to GPU while keep cold experts on CXL memory; 3) adaptive precision allocation on CXL-side experts to improve NDP performance.}
  \label{fig:design}
\end{figure*}

However, efficiently deploying MoE on \emph{GPU-NDP} systems remains challenging. Prior GPU-NDP MoE systems are largely \emph{context-agnostic} and rely on reactive or static policies that ignore the inherent dynamism of MoE routing: expert activation varies across layers, decoding steps, and even input sequences. As a result, on-demand expert placement can trigger unnecessary migrations between the GPU and the CXL-NDP tier, causing bandwidth contention. Static expert placement also presents a problem: experts mapped to NDP may suddenly become frequently activated (hot) and impose heavy compute pressure, while GPU-resident experts may become rarely activated (cold) and remain underutilized. Moreover, NDP compute units operate under tight power and area budgets, and even executing cold experts at full precision can introduce significant compute pressure and erode the benefits of near-data execution. shift the bottleneck to the NDP side, and erode the benefits of near-data execution.

To this end, we introduce a \emph{context-aware} expert placement and quantization strategy for efficient MoE inference on GPU–NDP system, as shown in Figure~\ref{fig:design}. Our design leverages runtime prefill statistics to guide both expert \emph{placement} and \emph{precision}. Our main contributions are:

\textbf{1) Empirical analysis of context-aware expert behavior.}  
We quantify the context dependence of MoE routing and show that expert activations vary significantly across decoding steps and input sequences, making static and on-demand expert placement ineffective. 

\textbf{2) Prefill-guided expert placement.}  
We further observe that prefill-stage routing distributions strongly predict decoding-stage behavior. This finding enables our informed expert placement: During the prefill stage of each sequence, we collect expert activation statistics to determine its importance. Important experts are placed on the GPU in full precision, while the remaining experts stay on NDP in low precision. The decoding stage then follows this prefill-guided placement, preserving MoE’s context awareness without incurring frequent expert migration.

\textbf{3) Context-aware mixed precision for NDP.}  
We adopt a mixed-precision quantization inspired by the recent method MC~\cite{huang2024mixture}. For each NDP-resident expert, we cache a set of GPTQ~\cite{frantar2022gptq}-quantized replicas at different precisions. We then apply a prefix-structured mixed-precision allocation to assign bitwidths based on the same prefill-stage expert importance information and a precomputed quantization loss table.
\section{Background and Motivation}\label{sec:bg}
% MoE Transformers are fundamentally \emph{memory-bound}: per-token routing induces small, rapidly shifting working sets, so fetching expert weights becomes the dominant bottleneck. 
% Coupling GPUs with CXL-NDP can mitigate this by executing \emph{cold} experts in place. However, prior systems remain context-agnostic and rely on reactive, on-demand expert placement policies. Post-training quantization further relieves NDP’s limited compute capacity, but pushing precision below 4 bits incurs non-negligible accuracy loss when applied uniformly across experts. We suggest a \textbf{context-aware} approach that jointly manages expert placement and precision, allocating bits and residency based on runtime activation statistics to preserve accuracy while maximizing efficiency.

In this section, we first outline the MoE Transformer architecture and explain why its routing dynamics make inference highly memory-bound. We then describe GPU-NDP hybrid systems, highlighting how near-data execution mitigates weight-transfer overheads but introduces new challenges due to context-agnostic expert placement. Finally, we discuss quantization for MoE models and motivate the need for a context-aware strategy to match NDP constraints while preserving accuracy.

\subsection{MoE-based Transformers}
In MoE-based Transformers, each feed-forward network (FFN) is replaced by a set of expert FFNs, and a router selects a small subset per token. 
Given a hidden state $\mathbf{x} \in \mathbb{R}^{d}$, the router $R(\mathbf{x})$ produces scores $w = \text{Softmax}(W_g \mathbf{x})$, and only the top-$k$ experts are activated. Each expert FFN typically consists of two linear layers with an intermediate activation.
% \[
% \text{MoE}(\mathbf{x}) = \sum_{i \in \text{Top-}k(w)} w_i \, E_i(\mathbf{x}),
% \]
% where $E_i(\cdot)$ denotes the $i$-th expert FFN (typically two linear layers with an intermediate activation).

% In practice, due to the massive parameter size of MoE models, they are often implemented using an \textit{offloading} mechanism, where expert parameters are stored in secondary memory and transferred to GPU for computation on demand~\cite{kim2024monde, tang2024hobbit, kamahori2024fiddler}. However, since the router dynamically determines expert selection at runtime, each token requires fetching a different subset of experts, especially during the decoding stage. This behavior causes frequent data transfers between the GPU and secondary memory, typically over the external memory interface. As shown in Figure~\ref{fig:transfer_ratio}, the data transfer time dominates the overall inference latency, while expert computation occupies only a small portion. Therefore, MoE inference under the offloading scenario is typically \textit{memory-bound}.

In practice, the parameter footprint of MoE models exceeds on-package HBM capacity, so expert parameters are placed in an external tier and fetched on demand during inference~\cite{kim2024monde,tang2024hobbit,kamahori2024fiddler}. 
% In this work, the external tier is \emph{CXL-attached memory with near-data processing (NDP)}. 
The router’s per-token, per-layer decisions yield small and rapidly changing working sets, which makes naive weight fetching particularly costly during the decoding stage.

\subsection{GPU--NDP Hybrid Systems for MoE}

To address the memory-bound nature of MoE inference and the limited capacity of GPU HBM, recent work has explored heterogeneous systems that couple GPUs with near-data processing (NDP) devices. Among these, CXL-attached NDP devices provide large-capacity DDR-class memory and high internal bandwidth. They can execute computations adjacent to offloaded parameters and support much larger MoE models at lower cost, making them a practical and deployable solution.

\begin{figure} % r=右侧，l=左侧，宽度自调
    \vspace{-10pt}
    \centering
   \includegraphics[width=0.45\textwidth]{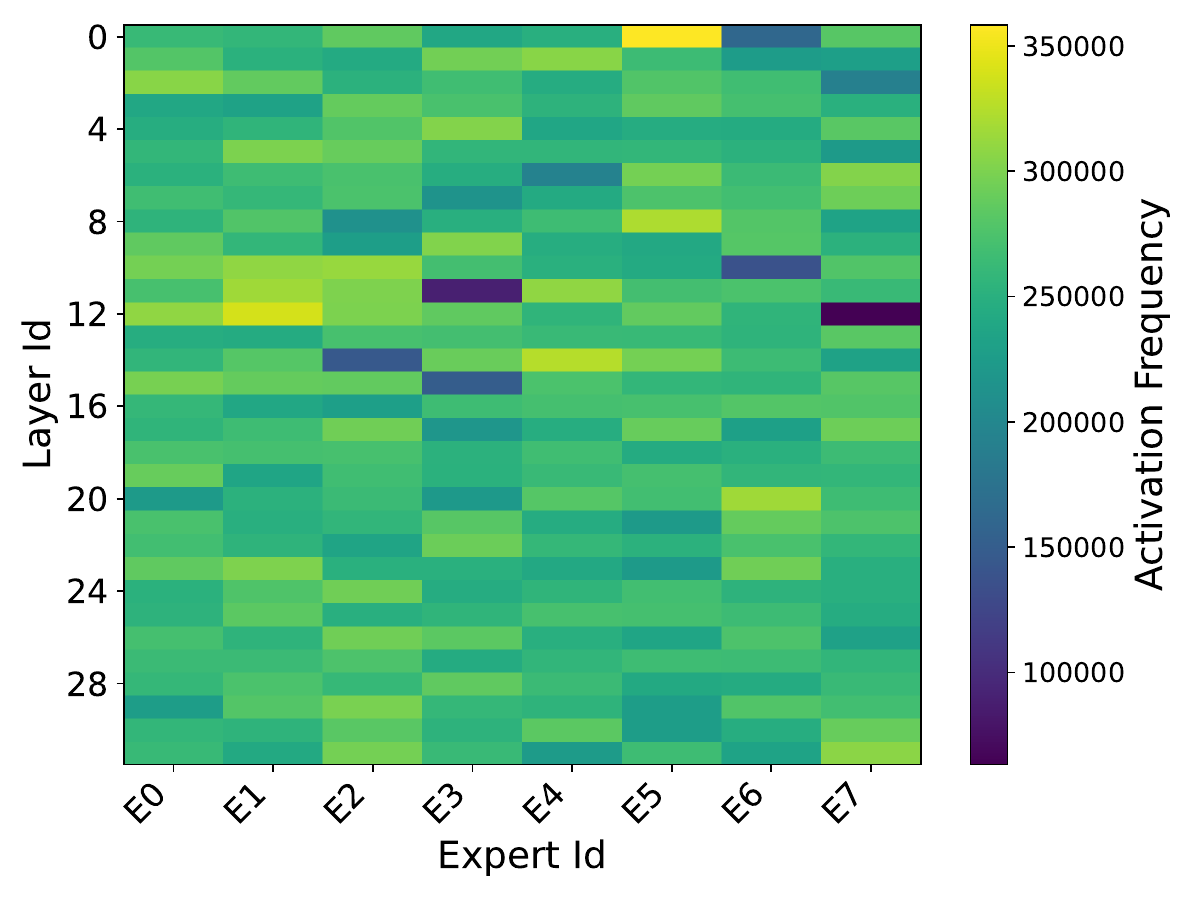}
  \caption{Activation frequency of all experts in Mixtral-8x7B.}
  \label{fig:expert_activation_distribution}
    \vspace{-5pt}
\end{figure}

Building on the observation that expert activations are highly skewed, recent GPU-NDP MoE systems such as MoNDE~\cite{kim2024monde} and PIMoE~\cite{wu2025pimoe} introduce the concept of \textit{hot} and \textit{cold} experts. As shown in Figure~\ref{fig:expert_activation_distribution}, which reports the activation frequency of all experts in Mixtral-8$\times$7B~\cite{jiang2024mixtral} on the WikiText-2~\cite{merity2016pointer} task, the distribution is far from uniform: a few experts are frequently activated, whereas some remain rarely used. This skew implies heterogeneous arithmetic intensities (compute-to-memory ratios) across experts, suggesting a device-aware mapping: pin hot, compute-intensive experts in GPU HBM, and place cold or infrequently used experts in the NDP tier, effectively turning \emph{parameter movement} into cheaper \emph{activation movement}~\cite{kim2024monde}.

However, prior GPU–NDP MoE systems largely rely on \emph{on-demand} swapping and \emph{context-agnostic} decisions at inference time. Under limited GPU$\leftrightarrow$CXL memory bandwidth, such reactive policies still incur substantial expert-transfer overheads and can reduce GPU utilization, which limits their efficiency and fails to fully exploit the inherent hot–cold characteristics of MoE experts.
% These observations motivate a context-aware runtime that collects expert activations, proactively manages expert placement across GPU and CXL memory, and use CXL-NDP to process activated experts outside of GPU.

% PIM devices provide much higher internal bandwidth and enable in-memory computation, which can effectively reduce data transfers of offloaded experts. Recent GPU--PIM MoE systems such as MoNDE~\cite{kim2024monde} and PIMoE~\cite{wu2025pimoe} introduce the concept of \textit{hot} and \textit{cold} experts. As shown in Figure~\ref{fig:expert_activation_distribution}, which reports the activation frequency of all experts in Mixtral-8×7B~\cite{} on the WikiText-2~\cite{} task, the activation is not uniform: a few experts are frequently activated, whereas some remain rarely used. This imbalance implies diverse operational intensities (i.e., compute-to-memory ratios) among experts, making it suitable to execute different experts on different devices: GPUs for compute-intensive experts and PIM for memory-intensive or infrequently used ones.  
% However, both MoNDE and PIMoE rely on \textit{on-demand} expert swapping during inference and are context-unaware. 
% Such strategy still incurs substantial expert transfer overheads, which limits their efficiency and fails to fully exploit the inherent hot–cold characteristics of MoE experts. 

% \begin{figure}[!ht]
%   \centering
%   \includegraphics[width=0.4\textwidth]{Figure/plot2_wiki.pdf}
%   \caption{Activation frequency of all experts in Mixtral-8x7B.}
%   \label{fig:expert_activation_distribution}
% \end{figure}

\subsection{Quantization for MoE Models}

To prevent the NDP tier from becoming the bottleneck in our GPU and CXL-NDP pipeline, we quantize cold experts executed on NDP. Unlike GPUs, NDP devices operate under tight power and area budgets and offer limited compute throughput and scratchpad capacity. Reducing arithmetic precision can increase effective in-device bandwidth and parallelism, while also lowering the energy per operation.

Accordingly, we apply post-training quantization (PTQ) \cite{xiao2023smoothquant, dettmers2022gpt3, lin2024awq, shao2023omniquant, frantar2022gptq, badri2023hqq} methods such as GPTQ~\cite{frantar2022gptq} to NDP-side experts. These techniques are practical for weight-only compression and can reliably achieve 4-bit precision. However, uniformly reducing expert precision below 4 bits leads to non-negligible accuracy loss. We attribute this degradation to overlooking the heterogeneous \emph{importance} and \emph{sensitivity} of experts, as well as their non-uniform activation behavior. We thus adopt a \textbf{context-aware} quantization strategy that adapts expert precision based on runtime activation statistics. This design matches NDP’s performance constraints while preserving accuracy, and it will be detailed in Section~\ref{sec:bitwidth-selector}.

\section{Key Observations on Context Awareness}\label{sec:moti}

% In this section, we analyze the key challenges that motivate our context-aware GPU--PIM MoE design. 
% Specifically, we first present the dynamic nature of expert activations in MoE models, which makes static expert partitioning ineffective. 
% We then show that the prefill-stage expert activation probability distribution is highly similar to that of the decoding stage and can be used to guide expert placement for the remainder of the inference, enabling context-aware expert placement without incurring frequent migrations.
% We finally discuss the limited computational capacity of current PIM devices, which constrains the performance of full-precision expert computation. 
% Together, these observations motivate a new system design that adaptively adjusts expert placement and precision according to runtime context.

We make two observations about the context dependence of MoE inference that motivate context-aware methods on a GPU--NDP system. (i) Expert activations vary across requests and steps, making static expert partitioning ineffective. (ii) The prefill-stage routing distribution is a strong early indicator of decoding-stage behavior, enabling proactive expert placement with minimal migrations. Guided by these, we develop a context-aware method that dynamically pins hot experts in GPU HBM while executing cold experts in-place on NDP, and adapts per-expert bitwidth based on runtime activation statistics to improve performance.

\subsection{Context-Dependent Expert Activations}
\label{sec:dynamic}

\begin{figure}[!ht]
    \centering
    \includegraphics[width=\linewidth]{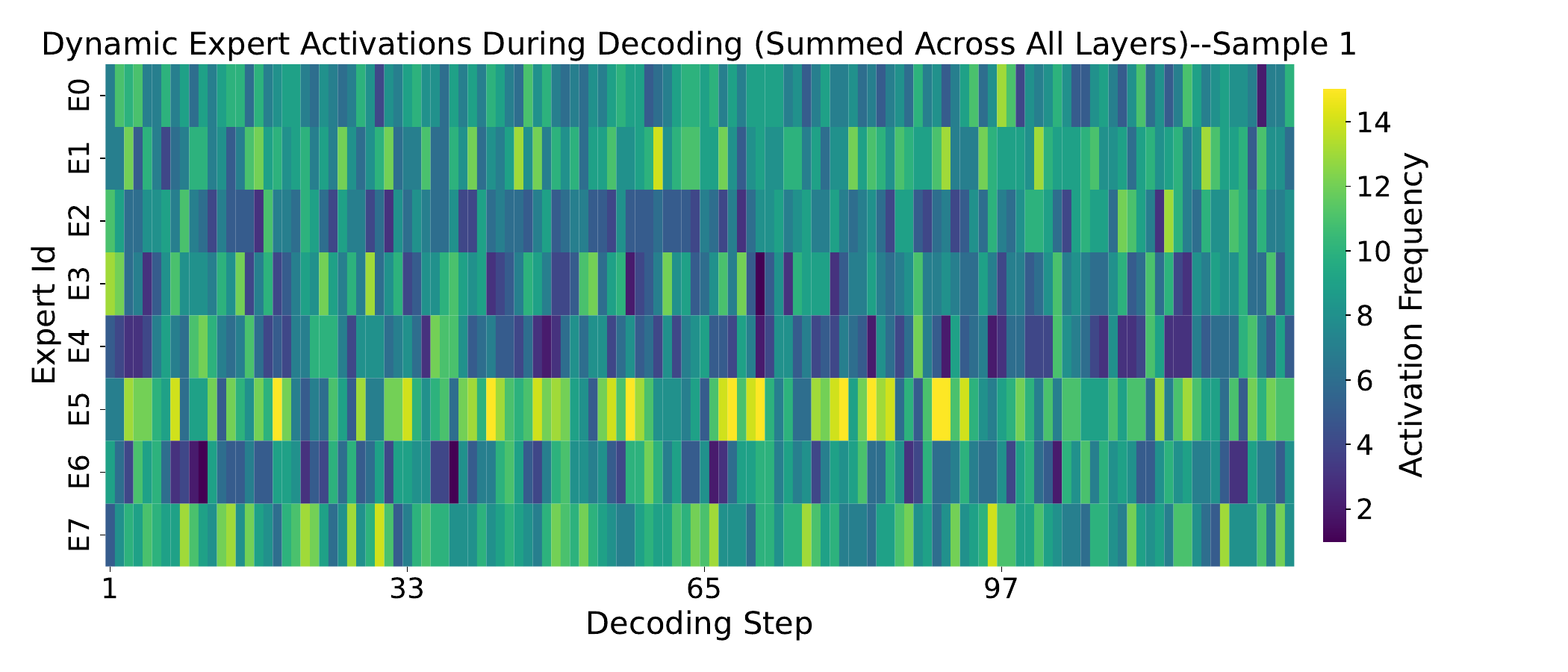}

    % 两张图之间如果想稍微拉开一点，可以手动加一点竖直间距
    \vspace{4pt}

    \includegraphics[width=\linewidth]{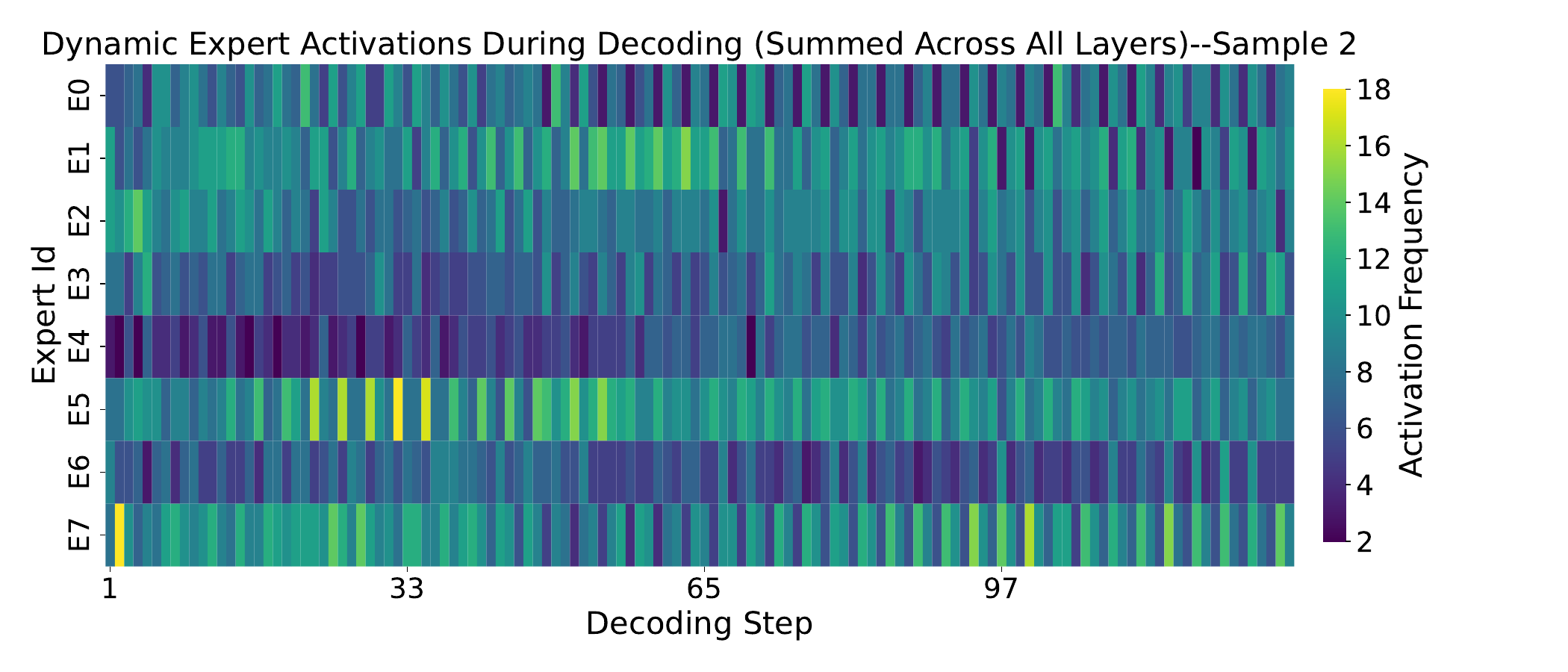}

    \caption{Different expert activation patterns for two samples with Mixtral-8$\times$7B on the C4 dataset, indicating the context dependence.}
    \label{fig:expert_dynamics}
\end{figure}

MoE models exhibit highly dynamic expert activation patterns. 
% During inference, the router determines the active experts for each token based on the current hidden representation, resulting in strong input- and context-dependence. 
Our empirical observations show that, during the inference, the distribution of activated experts changes significantly between consecutive decoding steps and even across different inputs. 
Figure~\ref{fig:expert_dynamics} illustrates these variations using two randomly sampled inputs from the C4~\cite{raffel2020exploring} dataset. In Sample 1, the activated experts exhibit highly irregular behavior across decoding steps, with neither uniform activation patterns nor similarity between adjacent steps. In Sample 2, the activation pattern across decoding steps appears more structured. However, this contrast between the two samples highlights a key phenomenon: expert activation is highly context-dependent. Even within the same dataset, different input sequences can trigger different activation behaviors.

% MoE routers select top-$k$ experts per token from the current hidden state, yielding input- and context-dependent activation patterns. Empirically, the activated-expert distribution shifts across requests, layers, and even consecutive decode steps; Figure~\ref{fig:expert_dynamics} shows two C4 samples with different activation patterns. In Sample 1, the activated experts exhibit highly irregular behavior across decoding steps, with neither uniform activation patterns nor similarity between adjacent steps. In Sample 2, the activation pattern across decoding steps appears more structured. Hence, an expert that is ``cold'' on average may become transiently ``hot'' for specific contexts.

Such variability implies that a static placement or a global frequency metric cannot capture an expert’s importance. Therefore, merely assigning experts to different devices is insufficient, and the placement should be dynamically adjusted based on contexts.

\subsection{Context-Aware Opportunities from Prefill}
\label{sec:context-aware opportunities}
While expert placement in GPU–NDP systems should ideally be dynamic, this dynamism also brings a practical challenge. The purpose of introducing NDP is to reduce the overhead caused by expert offloading and migration. If the system updates the placement too frequently, for instance at every decoding step, the additional transfers may eliminate the bandwidth benefits provided by NDP. As a result, it becomes important to define appropriate conditions and timing for expert migration.

\begin{figure}[!ht]
        \centering
        \includegraphics[width=0.4\textwidth]{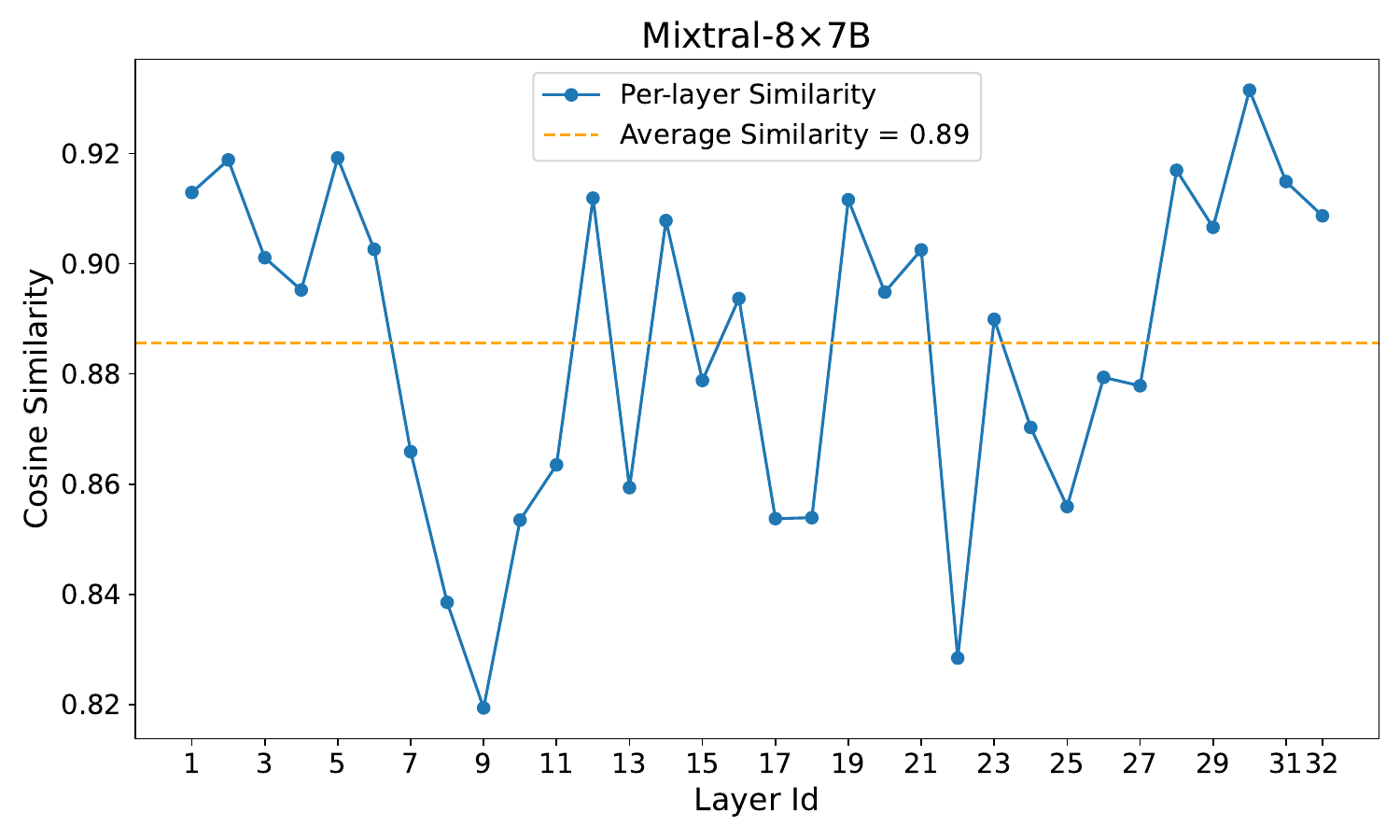}
    \caption{Expert activation similarities between prefill and decoding, motivating context-aware design.}
    \label{fig:sim_truthfulqa}
\end{figure}

Fortunately, our analysis provides a useful observation that helps address this issue. Within the same sequence, the expert activation probability distribution during the prefill stage is often very similar to the distribution observed during decoding. Figure~\ref{fig:sim_truthfulqa} shows this effect for the Mixtral-8×7B~\cite{jiang2024mixtral} model on the TruthfulQA~\cite{lin2022truthfulqa} task. We compute the cosine similarity between the prefill and decoding expert activation probability distributions and report the average across all samples. Mixtral-8×7B has eight experts per layer, and the average similarity across all layers reaches 0.89.

These results indicate that the prefill stage already provides a reliable estimate of how experts will be activated during decoding. Therefore, the activation statistics collected in the prefill stage can be used to guide expert placement for the remainder of the inference. This approach helps avoid unnecessary migrations while still capturing context-aware activation behavior.

\begin{algorithm}[t]
\caption{Context-aware Expert Placement and Quantization}
\label{alg:context-aware-moe}
\begin{algorithmic}[1]
\Require MoE model with $L$ layers, $E$ experts; GPU-side expert budget $K$ (per layer); NDP-side expert avg.\ bitwidth $\bar b$; mixing coefficient $\alpha$; calibration dataset $\mathcal{D}_{\mathrm{cal}}$
\Statex

\State \textbf{Offline Calibration (once)}
\For{$l=1$ to $L$}
  \For{$e=1$ to $E$}
    \For{$b \in \{1,2,3,4\}$}
      \State Estimate loss $L_{l,e}(b)$ on $\mathcal{D}_{\mathrm{cal}}$
    \EndFor
  \EndFor
\EndFor

\Statex
\State \textbf{Online Inference (for each sequence)}
\State \textbf{Prefill:}
\State Run prefill and collect for each layer $l$ and expert $e$: activation counts $P_{l,e}$ and routing-score sums $W_{l,e}$.
\State \textbf{Expert Importance and Placement}
\For{$l=1$ to $L$}
  \State $\tilde P_{l,e} \gets \mathrm{Norm}(P_{l,e}),\;\tilde W_{l,e} \gets \mathrm{Norm}(W_{l,e})$
  \State $S_{l,e} \gets \alpha \tilde P_{l,e} + (1{-}\alpha)\tilde W_{l,e}$
  \State $\mathcal{H}_l \gets$ top-$K$ experts by $S_{l,e}$ \Comment{GPU, FP16}
  \State $\mathcal{C}_l \gets \{e \mid e \notin \mathcal{H}_l\}$ \Comment{experts on NDP}
\EndFor
\State \textbf{Expert Bitwidth Assignment on NDP (Prefix-Split)}
\For{$l=1$ to $L$}
  \State $b_{l,e} \gets \textsc{PrefixSplit}(\{S_{l,e}\}_{e\in\mathcal{C}_l}, \{L_{l,e}(b)\}_{e\in\mathcal{C}_l}, \bar b)$ \Comment{Sec.\ref{sec:bitwidth-selector}}
\EndFor
\State \textbf{Decoding:}
\For{each decoding step}
  \For{each selected expert $e$ in layer $l$}
    \If{$e\in\mathcal{H}_l$} \State Run expert on GPU (FP16)
    \Else \State Run expert on NDP with bitwidth $b_{l,e}$ \EndIf
  \EndFor
\EndFor

\end{algorithmic}
\end{algorithm}

\section{Context-Aware MoE System Design}\label{sec:design}

This section presents our context-aware MoE system design based on GPU--NDP, which consists of two tightly coupled components, as shown in Figure~\ref{fig:design} : (i) a \textit{dynamic expert placement module} that leverages routing statistics collected during the prefill stage to decide which experts reside on GPU and which remain on CXL-NDP, and (ii) a \textit{dynamic bit-width selector} that applies mixed-precision quantization to NDP-resident experts under a per-layer bit-width budget. Together, these components exploit the contextual activation dynamics of MoE models and enable efficient inference with minimal expert migration.

\subsection{Expert Placement Module}

As shown in Section~\ref{sec:context-aware opportunities}, for the same input sequence, the expert activation distributions from prefill closely match those in decoding. We leverage this property to determine the GPU--NDP expert placement \textit{once per sequence}. As shown in Lines 9–18 of Algorithm~\ref{alg:context-aware-moe}, at the beginning of prefill, we collect two statistics for each expert $e$ in each MoE layer $l$: (i) its activation frequency $P_{l,e}$ and (ii) its accumulated routing score $W_{l,e}$. These two metrics capture different aspects of expert usage: frequency reflects how often an expert is selected, while routing score reflects the confidence of each activation. Then we compute a normalized importance score:
\[
S_{l,e} \;=\; \alpha\,\widetilde{P}_{l,e} \;+\; (1-\alpha)\,\widetilde{W}_{l,e},
\]
where $\alpha \in [0,1]$ controls the tradeoff and $\widetilde{P}_{l,e}$ and $\widetilde{W}_{l,e}$ denote normalized quantities. Based on $S_{l,e}$, we select the top-$K$ most important experts in each layer and migrate them to GPU in FP16, subject to the GPU memory capacity. The remaining experts reside on CXL-NDP. Importantly, expert placement is performed \textit{only once} after the prefill stage, and the decoding stage uses this fixed placement without further migration, i.e., each sequence undergoes only a single expert migration. For each routing decision, the computation is performed on the device hosts the selected experts.

\subsection{Expert Bitwidth Selector on NDP}
\label{sec:bitwidth-selector}

After expert placement is fixed, we quantize the experts that remain on NDP to reduce compute pressure during inference. Rather than quantizing weights on the fly, for each NDP-resident expert we cache a set of GPTQ~\cite{frantar2022gptq}-quantized replicas at bitwidths ${1,2,3,4}$. We then use a prefix-structured mixed-precision allocation to assign bitwidths to NDP-resident experts under a layer-wise average bitwidth budget, as shown in Lines 1–8 and 19–22 of Algorithm~\ref{alg:context-aware-moe}. Consider a layer with $E$ total experts, among which $E_{\mathrm{NDP}}$ are placed on NDP. For these $E_{\mathrm{NDP}}$ NDP-resident experts, we allow candidate bitwidths $\{1,2,3,4\}$, taking 1-bit as the initial bitwidth. This discrete bitwidth set enables average 2-bit or 3-bit quantization by mixing experts assigned to different bitwidths. Let $\bar b$ denote the target average bitwidth on NDP for this layer. The corresponding number of ``bitwidth increments'' is
$
R=E_{\mathrm{NDP}}(\bar b - 1),
$
where each step $1\!\to\!2$, $2\!\to\!3$, or $3\!\to\!4$ consumes one unit of this increment budget.
We reuse the importance ordering obtained in the expert placement module and index experts in descending order of importance as $i=1,\dots,E$. We construct a per-layer loss table using a calibration dataset such as C4. For each expert $i$ and each bitwidth $b \in \{1,2,3,4\}$, we measure a loss $L_i(b)$, defined as the MSE between the quantized output and a full-precision reference. From this table we extract the direct gains of moving from 1-bit to higher precisions:
\[
\Delta_i(2) = L_i(1)-L_i(2),\quad
\Delta_i(3) = L_i(1)-L_i(3),
\]
\[
\Delta_i(4) = L_i(1)-L_i(4).
\]
To evaluate the total gain achieved by assigning higher bitwidths to the more important experts, we first accumulate these per-expert gains into prefix sums. Specifically, for any $k$, the cumulative benefits of upgrading the top-$k$ experts from 1-bit to 2-, 3-, or 4-bit are
\[
C_{2}(k) = \sum_{i=1}^{k} \Delta_i(2),\quad
C_{3}(k) = \sum_{i=1}^{k} \Delta_i(3),\quad
C_{4}(k) = \sum_{i=1}^{k} \Delta_i(4).
\]
To keep the assignment aligned with the importance ranking, we enforce a prefix structure: more important experts receive quantization configurations with smaller expected loss (usually larger bitwidths), and less important ones receive progressively coarser settings. Let $n_4,n_3,n_2,n_1$ be the numbers of experts assigned to $4$-, $3$-, $2$-, and $1$-bit, respectively, with
\[
n_4 + n_3 + n_2 + n_1 = E_{\mathrm{NDP}}.
\]
By treating 1-bit as the initial bitwidth, the bitwidth budget translates to
\[
3n_4 + 2n_3 + n_2 = R.
\]
Given this structure, the total gain achieved by assigning $(n_4,n_3,n_2)$ higher-precision experts is
\begin{align*}
G(n_4,n_3,n_2)
&= C_{4}(n_4) \\
&\quad + \big[C_{3}(n_4{+}n_3)-C_{3}(n_4)\big] \\
&\quad + \big[C_{2}(n_4{+}n_3{+}n_2)-C_{2}(n_4{+}n_3)\big].
\end{align*}

Intuitively, once the experts are sorted by importance, a prefix-structured assignment is fully determined by how many of the most important experts use each bitwidth. We therefore treat the counts $(n_4,n_3)$ as search variables and derive $n_2$ and $n_1$ from the constraints. Concretely, we let $n_4$ range over all integers such that $0 \le n_4 \le E_{\mathrm{NDP}}$ and $3n_4 \le R$, and for each fixed $n_4$ we let $n_3$ range over all integers such that $0 \le n_3 \le E_{\mathrm{NDP}}-n_4$ and $3n_4 + 2n_3 \le R$. For each feasible $(n_4,n_3,n_2)$, the total gain $G(n_4,n_3,n_2)$ can be evaluated in constant time using the prefix sums $C_2(\cdot),C_3(\cdot),C_4(\cdot)$, and we simply keep the tuple $(n_4^\star,n_3^\star,n_2^\star)$ that attains the largest gain. Since the dominant time cost comes from enumerating all feasible $(n_4,n_3)$ pairs, the per-layer time complexity is $O(E_{\mathrm{NDP}}^2)$. And as we only apply this selection to NDP-resident experts and $E_{\mathrm{NDP}}$ is not large in practice, the overall overhead $O(LE_{\mathrm{NDP}}^2)$ across $L$ layers is negligible relative to the inference cost.

Finally, once expert placement and NDP-side expert quantization bitwidth selection are completed before the decoding stage, we keep these configurations fixed during decoding, as shown in Lines 23–32 of Algorithm~\ref{alg:context-aware-moe}.
\section{Evaluation}\label{sec:eval}
% \textcolor{red}{
% \begin{itemize}
%     \item \textbf{Figure/Table 8:} Accuracy comparison: prefill tasks (\texttt{lm-harness}) and decoding tasks (\texttt{gsm8k}, \texttt{truthfulqa\_gen}); compare our dynamic bitwidth quantization method vs.\ MoNDE (lossless) vs.\ GPTQ uniform quantization.
%     \item \textbf{Figure 9:} Performance comparison (latency/throughput, and energy if time permits): Our Method vs.\ MoNDE vs.\ Hobbit (GPU only) on Mixtral-8$\times$7B and Mixtral-8$\times$22B.
%     \item \textbf{Figure 10:} PIM performance under different bitwidths: FP16 vs.\ our average bandwidth.
% \end{itemize}
% }

% In this section, we evaluate our method in terms of both end-to-end performance on a GPU–NDP system and model accuracy. We begin by describing the experimental setup in Section~\ref{sec:exp:setup}, including model configurations, datasets, system settings, and baseline methods. Since our method is primarily designed to reduce expert migration and NDP-side compute overhead to improve system efficiency, we first present end-to-end latency and throughput comparisons against both GPU–NDP and GPU-only baselines in Section~\ref{sec:exp:efficiency}. We then present accuracy results in Section~\ref{sec:exp:acc} to show that our method’s context-aware expert placement and bitwidth selector not only deliver significant performance gains but also provide additional benefits to model accuracy.

\subsection{Experimental Setup}
\label{sec:exp:setup}

\textit{Models and Datasets.} We evaluate two popular MoE models, Mixtral-8$\times$7B~\cite{jiang2024mixtral} and Mixtral-8$\times$22B~\cite{mistral2024mixtral}, to assess the effectiveness of our method on large-scale MoE models, as detailed in Table~\ref{tab:model-config}. We evaluate our method on a set of language understanding and reasoning benchmarks, including MMLU\cite{hendrycks2020measuring}, MathQA\cite{amini2019mathqa}, HellaSwag\cite{zellers2019hellaswag}, ARC-Easy\cite{clark2018think}, ARC-Challenge\cite{clark2018think}, BoolQ\cite{clark2019boolq}, WinoGrande\cite{sakaguchi2021winogrande}, and PIQA\cite{bisk2020piqa}. MMLU is evaluated in a 5-shot setting, while all other tasks use zero-shot evaluation. All accuracy results are obtained using the EleutherAI LM Evaluation Harness~\cite{eval-harness}. 

\begin{table}[ht]
\centering
\caption{Configs of evaluated MoE models}
\label{tab:model-config}
\resizebox{\columnwidth}{!}{%
\begin{tabular}{|l|c|c|c|c|c|c|}
\hline
\textbf{Model} & \textbf{Hidden} & \textbf{Layers} & \textbf{Experts} & \textbf{Top-k} & \textbf{Experts Params.} & \textbf{Params.} \\ \hline
Mixtral-8$\times$7B \cite{jiang2024mixtral}   & (4096, 14336)  & 32 & 8  & 2 & 45.1B  & 46.7B  \\ 
Mixtral-8$\times$22B \cite{mistral2024mixtral}  & (6144, 16384)  & 56 & 8  & 2 & 135.5B & 140.6B \\ \hline
\end{tabular}}
\\[0.25em]
\begin{minipage}{\columnwidth}
\footnotesize
\end{minipage}
\vspace{-8pt}
\end{table}

\textit{System Settings.} To enable a fair and intuitive comparison, we adopt the same GPU–NDP system configuration used in MoNDE~\cite{kim2024monde} and build our NDP system simulator on Ramulator~\cite{kim2015ramulator} following the same methodology. The detailed configuration is summarized in Table~\ref{tab:env-config}. Our method migrates only a small number of high-importance experts to the GPU, while the remaining low-bitwidth experts are executed on the NDP devices. For the GPU-only baseline, all experts are loaded to the GPU on demand via PCIe. We evaluate multiple input–output length configurations to measure end-to-end latency and decoding throughput. Considering GPU memory capacity, for Mixtral-8$\times$7B we place four hot experts per layer on the GPU and the remaining four on the NDP device. For Mixtral-8$\times$22B, two hot experts per layer reside on the GPU and six are executed on the NDP device.

\begin{table}[h]
\centering
\caption{System configurations}
\resizebox{\columnwidth}{!}{
\begin{tabular}{|l|l|l|}
\hline
\textbf{Component} & \textbf{Specification} & \textbf{Value} \\ \hline

\multirow{2}{*}{System} 
       & Composition & 1$\times$H100 GPU + 1$\times$DDR-based NDP \\ \cline{2-3}
       & Interconnect & PCIe Gen4 $\times$16 \\ \hline

\multirow{3}{*}{GPU (H100)} 
       & SM Count & 132 \\ \cline{2-3}
       & Peak Compute & 989.4 TFLOP/s \\ \cline{2-3}
       & Memory Capacity & 80 GB HBM3 \\ \hline

\multirow{4}{*}{NDP Device} 
       & Capacity & 512 GB \\ \cline{2-3}
       & Bandwidth & 512 GB/s \\ \cline{2-3}
       & Compute Units & 64 $\times$ (4$\times$4 Systolic Arrays) \\ \cline{2-3}
       & Clock Frequency & 1 GHz \\ \hline

\end{tabular}}
\label{tab:env-config}
\end{table}

\textit{Baselines.} For performance evaluation, we use MoNDE~\cite{kim2024monde} on the same GPU–NDP system as our primary baseline. We also include a GPU-only baseline HOBBIT~\cite{tang2024hobbit}, which is a mixed-precision expert offloading system. For accuracy evaluation, we mainly compare our method against the accuracy of the original full-precision models, which also represents the accuracy of MoNDE.

\subsection{Performance Evaluation}
\label{sec:exp:efficiency}

\begin{figure}[htbp]
        \centering
        \includegraphics[width=0.45\textwidth]{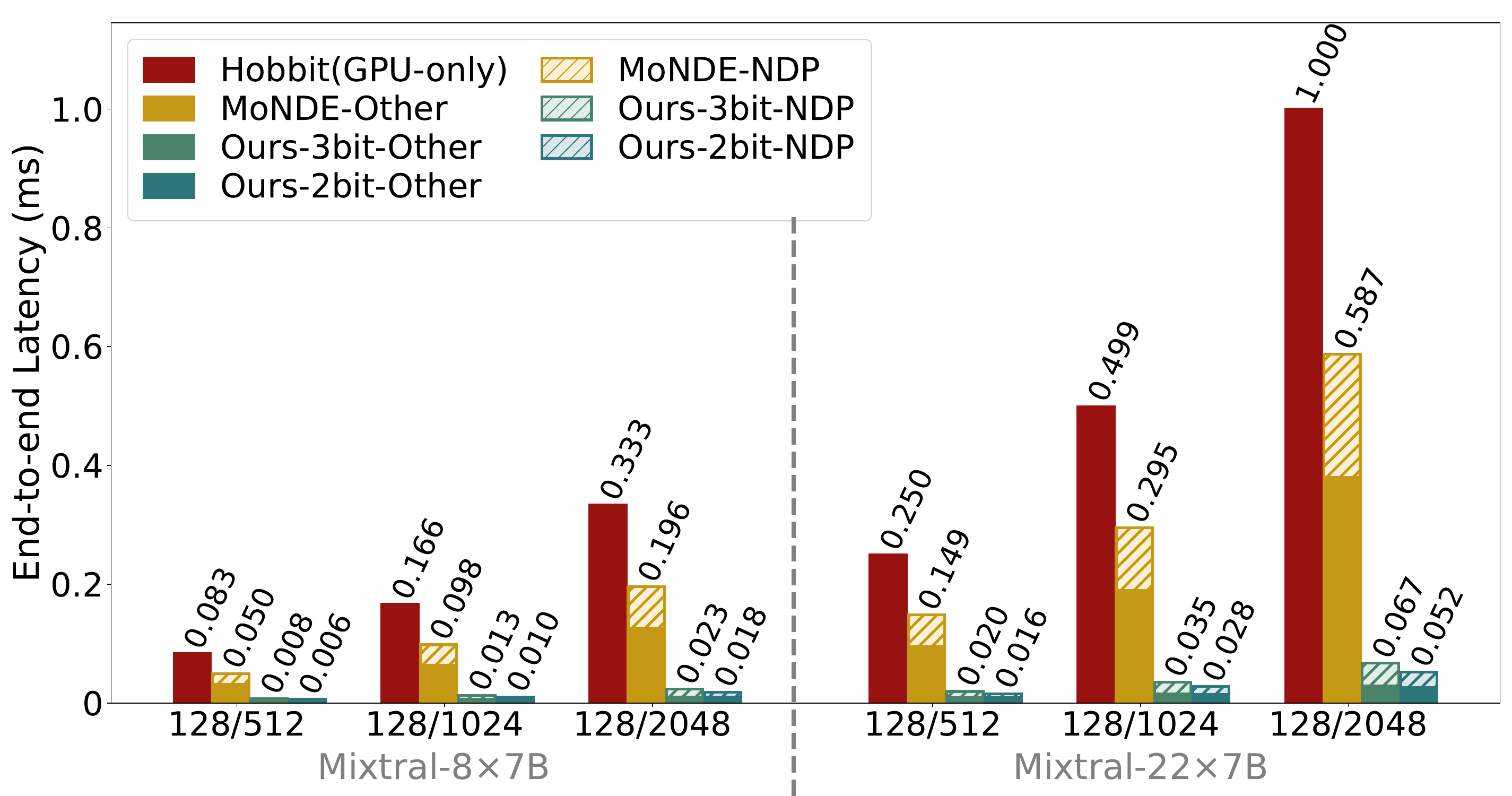}
    \caption{End-to-end latency comparison across different methods, with NDP-side latency shown separately to highlight the benefits of our method in both reducing NDP computation and minimizing expert migration.}
    \label{fig:lat}
\end{figure}

\begin{figure}[htbp]
        \centering
        \includegraphics[width=0.45\textwidth]{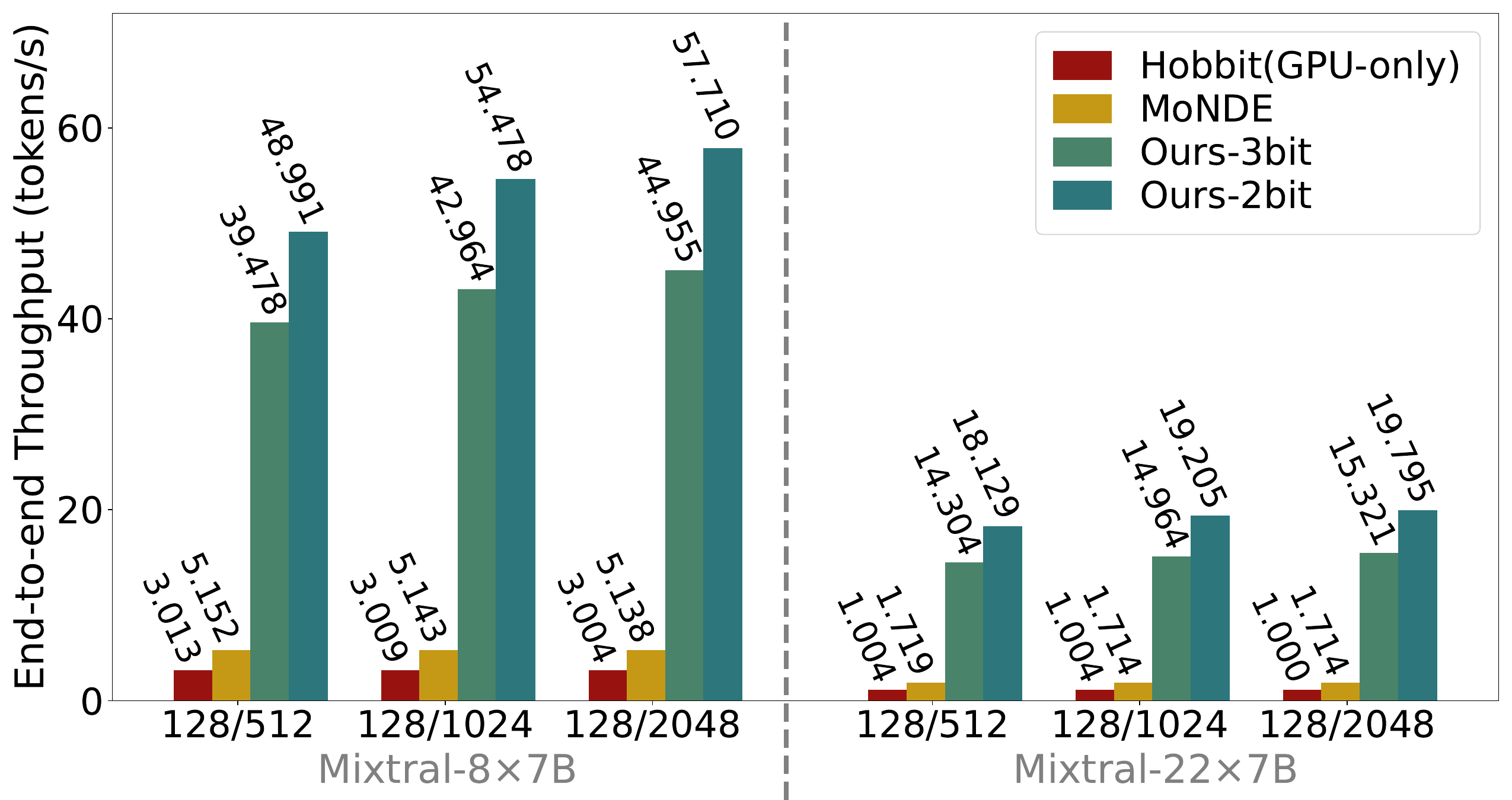}
    \caption{End-to-end decoding throughput.}
    \label{fig:thr}
    \vspace{-10pt}
\end{figure}

We evaluate end-to-end latency and decoding throughput on two large MoE models, Mixtral-8$\times$7B and Mixtral-8$\times$22B, to demonstrate the benefits of the GPU--NDP system. Experiments are conducted under multiple input/output length configurations, and we additionally report the isolated NDP-side latency improvements. Figures~\ref{fig:lat} and~\ref{fig:thr} compare our method with all baselines. ``Ours-3bit'' and ``Ours-2bit'' denote average 3-bit and 2-bit bitwidth on the NDP device.

On Mixtral-8$\times$7B, Ours-3bit achieves a \textbf{6.6--8.3$\times$} end-to-end speedup over MoNDE on the same GPU--NDP system, while Ours-2bit achieves \textbf{7.9--10.6$\times$}. The corresponding decoding throughput improvements reach \textbf{8.7$\times$} and \textbf{11.2$\times$}, respectively. On Mixtral-8$\times$22B, Ours-3bit attains \textbf{7.6--8.7$\times$} speedup and Ours-2bit achieves \textbf{9.5--11.2$\times$}, with decoding throughput gains of \textbf{8.9$\times$} and \textbf{11.5$\times$}. For both models, the NDP-side execution alone sees approximately \textbf{5$\times$} (3-bit) and \textbf{8$\times$} (2-bit) latency reductions.

These results show that our method benefits not only from low-precision NDP execution but, more importantly, from significant reductions in expert migration, which improves overall pipeline efficiency. Compared with the GPU-only baseline (Hobbit), our method delivers even larger gains: Ours-2bit achieves up to \textbf{18$\times$} speedup on Mixtral-8$\times$7B and \textbf{19$\times$} on Mixtral-8$\times$22B.

\subsection{Accuracy Evaluation}
\label{sec:exp:acc}

We conduct accuracy evaluation on Mixtral-8$\times$7B and compare our method against two baselines. The first is MoNDE, whose experts are computed in full precision and therefore provide lossless accuracy. The second is a our method variant without the Expert Bitwidth Selector, which allows us to isolate the benefit brought by context-aware bitwidth selection. For our method, we set the importance-score parameter $\alpha=0.5$ and use 1024 samples from the C4 dataset as the calibration set for constructing the loss table $L_{l,e}(b)$. As described in Section~\ref{sec:bitwidth-selector}, GPTQ is used as the underlying quantization method. In addition, consistent with the performance evaluation setup, we place four experts per layer on the GPU and the remaining four on the NDP device, which implies that half of the experts are executed in full precision (FP16).

% 导言区
% \usepackage{array}
\newcolumntype{L}[1]{>{\raggedright\arraybackslash}m{#1}}

\renewcommand{\arraystretch}{1.5}

\begin{table}[h]
\centering
\caption{Model accuracy comparison for Mixtral-8$\times$7B}
\resizebox{\columnwidth}{!}{
\begin{tabular}{|L{0.3\columnwidth}|c|c|c|c|c|c|c|c|c|}
\hline
\textbf{Methods} & \textbf{MMLU} & \textbf{MathQA} & \textbf{HellaSwag} & \textbf{ARC-E} & \textbf{ARC-C} & \textbf{BoolQ} & \textbf{WinoGrande} & \textbf{PIQA} & \textbf{Avg.} \\ \hline

Original (MoNDE) & 68.02 & 42.48 & 64.87 & 84.05 & 56.83 & 85.35 & 76.24 & 82.43 & 70.03 \\ \hline

Ours-3bit w/o Expert Bitwidth Selector
& 67.12 & 42.88 & 64.34 & 83.37 & 55.46 & 85.02 & 76.87 & 82.64 & 69.71 \\ \hline

Ours-3bit
& 66.72 & 42.81 & 64.08 & 83.82 & 55.70 & 85.57 & 77.38 & 83.12 & 69.90 \\ \hline

Ours-2bit w/o Expert Bitwidth Selector
& 56.64 & 34.41 & 59.11 & 79.16 & 50.00 & 78.75 & 70.95 & 78.78 & 63.48 \\ \hline

Ours-2bit
& 63.31 & 40.46 & 60.07 & 81.35 & 53.66 & 81.77 & 73.09 & 79.76 & 66.68 \\ \hline

\end{tabular}}
\label{tab:acc-4hot4cold}
\end{table}

Table~\ref{tab:acc-4hot4cold} reports the accuracy across multiple tasks. Ours-3bit incurs only a \textbf{0.13\%} average accuracy drop relative to the original model, while Ours-2bit shows an average drop of \textbf{3.4\%}. Compared with the variant without Expert Bitwidth Selector, Ours-3bit achieves a slight improvement and Ours-2bit yields a \textbf{3.2\%} average gain, demonstrating the effectiveness of the Expert Bitwidth Selector.
Overall, our method’s context-aware Expert Placement Module and Expert Bitwidth Selector significantly improve GPU--NDP performance while preserving accuracy close to that of the full-precision model.

\section{Related Work}\label{sec:related}

\textbf{Expert Offloading for MoE.}
Given the substantial GPU memory footprint of MoE-based LLMs, a large amount of recent work has explored how to efficiently offload experts to external memory. Several GPU-only systems improve offloading efficiency by prefetching, prediction, or expert scheduling, including Pre-gated MoE\cite{hwang2024pre}, eMoE\cite{tairin2025emoe}, MoE-Infinity\cite{xue2025moe}, SwapMoE\cite{kong2023swapmoe} and Klotski\cite{fang2025klotski}. Although HOBBIT\cite{tang2024hobbit} provides a GPU--CPU cooperative mode, its default and primary configuration executes both attention and expert FFNs on the GPU. These approaches indeed improve expert offloading efficiency, but they remain fundamentally constrained by the computational capacity of a GPU-only architecture and still suffer execution stalls. Other systems employ heterogeneous designs such as GPU--CPU architectures~\cite{zhong2025hybrimoe,kamahori2024fiddler,yuan2025moe,cao2025moe,zhang2025daop}, where low-workload or uncached experts, as well as attention layers, may be executed on the CPU. GPU--NDP architectures~\cite{kim2024monde,wu2025pimoe,yun2024duplex,huang2025hd} further extend this idea by enabling in-memory execution of experts and thereby avoiding frequent data transfers. Compared with CPUs, NDP devices provide significantly higher internal bandwidth and lower data-access latency, making them better suited for memory-intensive operations and full-parameter storage. However, existing GPU--NDP MoE systems are context-unaware and do not adapt expert execution to the computational characteristics of NDP, which leads to substantial expert migration overhead remaining unresolved.

\textbf{Quantization for MoE.}
Post-training quantization (PTQ)~\cite{dettmers2022gpt3,xiao2023smoothquant,shao2023omniquant,lin2024awq} is an effective technique for compressing LLMs without additional training. Methods such as GPTQ~\cite{frantar2022gptq} and HQQ~\cite{badri2023hqq} reduce model size by approximating weights with low-bit representations while preserving accuracy. Recent work extends PTQ to MoE models by quantizing experts to mitigate their large parameter footprint. MiLo\cite{huang2025milo} combines 3-bit quantization with low-rank compensators to recover accuracy, while MC\cite{huang2024mixture} proposes an expert-aware quantization method that automatically assigns an optimal bitwidth to each expert. In contrast to these goals, our method leverages expert quantization primarily to reduce NDP-side computational pressure rather than to minimize model storage.
\section{Conclusion}\label{sec:conclude}

In this work, we present a context-aware MoE inference system for hybrid \mbox{GPU--NDP} architectures based on CXL-attached memory devices. Our analysis establishes that prefill-stage routing reliably indicates decoding-time activations, enabling prefill-guided expert placement that dynamically pins hot experts in GPU HBM and executes cold experts in place on the NDP tier. To ensure our design falls into the limited compute capacity of NDP, we introduce context-aware mixed-precision quantization that allocates per-expert bitwidths, converting expensive parameter movement into cheaper activation movement and sustaining overlapped execution between GPU and NDP. 
% Experiments demonstrate reduced memory traffic and end-to-end latency, higher throughput and GPU utilization, and preserved quality over reactive offloading and uniform-precision baselines. 
Experiments demonstrate that our approach achieves up to 6.6–8.3× end-to-end speedup and 8.7× decoding throughput improvement over state-of-the-art method, while incurring only a 0.13\% average accuracy drop.

% Future work could include extending the runtime to multi-GPU and multi-NDP topologies and emerging CXL~3.x peer-to-peer paths, co-optimizing routing with residency and precision during training, and exploring NDP-friendly expert architectures and QoS-aware policies for service-level guarantees. \todo{optional}

%%
%% The next two lines define the bibliography style to be used, and
%% the bibliography file.
% \newpage
\bibliographystyle{ACM-Reference-Format}
\bibliography{refs}

\end{document}